\documentclass{article}

\usepackage{PRIMEarxiv}

\usepackage[utf8]{inputenc} % allow utf-8 input
\usepackage[T1]{fontenc}    % use 8-bit T1 fonts
\usepackage[colorlinks = true,
            linkcolor = green,
            urlcolor  = magenta,
            citecolor = blue,
            anchorcolor = black]{hyperref}
\usepackage{url}            % simple URL typesetting
\usepackage{booktabs}       % professional-quality tables
\usepackage{amsfonts}       % blackboard math symbols
\usepackage{nicefrac}       % compact symbols for 1/2, etc.
\usepackage{microtype}      % microtypography
\usepackage{fancyhdr}       % header
\usepackage{graphicx}       % graphics

\usepackage{amsmath}
\usepackage{amsthm}
\usepackage{authblk}

\usepackage[capitalize]{cleveref}
\crefname{section}{Sec.}{Secs.}
\Crefname{section}{Section}{Sections}
\Crefname{table}{Table}{Tables}
\crefname{table}{Tab.}{Tabs.}

\newcommand\blfootnote[1]{%
  \begingroup
  \renewcommand\thefootnote{}\footnote{#1}%
  \addtocounter{footnote}{-1}%
  \endgroup
}

\title{SwimXYZ: A large-scale dataset of synthetic swimming motions and videos}

\author[1]{Guénolé Fiche}
\author[2]{Vincent Sevestre $^\ast$}
\author[3]{Camila Gonzalez-Barral $^\ast$}
\author[1]{Simon Leglaive}
\author[1]{Renaud Séguier}
\affil[1]{CentraleSupélec, IETR UMR CNRS 6164, France}
\affil[2]{Centrale Nantes, France}
\affil[3]{Université Technologique de Compiègne, France}

\begin{document}
\maketitle
\blfootnote{$^\ast$ Work carried out at CentraleSupélec as part of an internship founded by the EUR DIGISPORT, project supported by the ANR within the framework of the PIA France 2030 (ANR-18-EURE-0022).}

\vspace{-2cm}
\begin{figure}[!h]
    \centering
    \includegraphics[width=\textwidth]{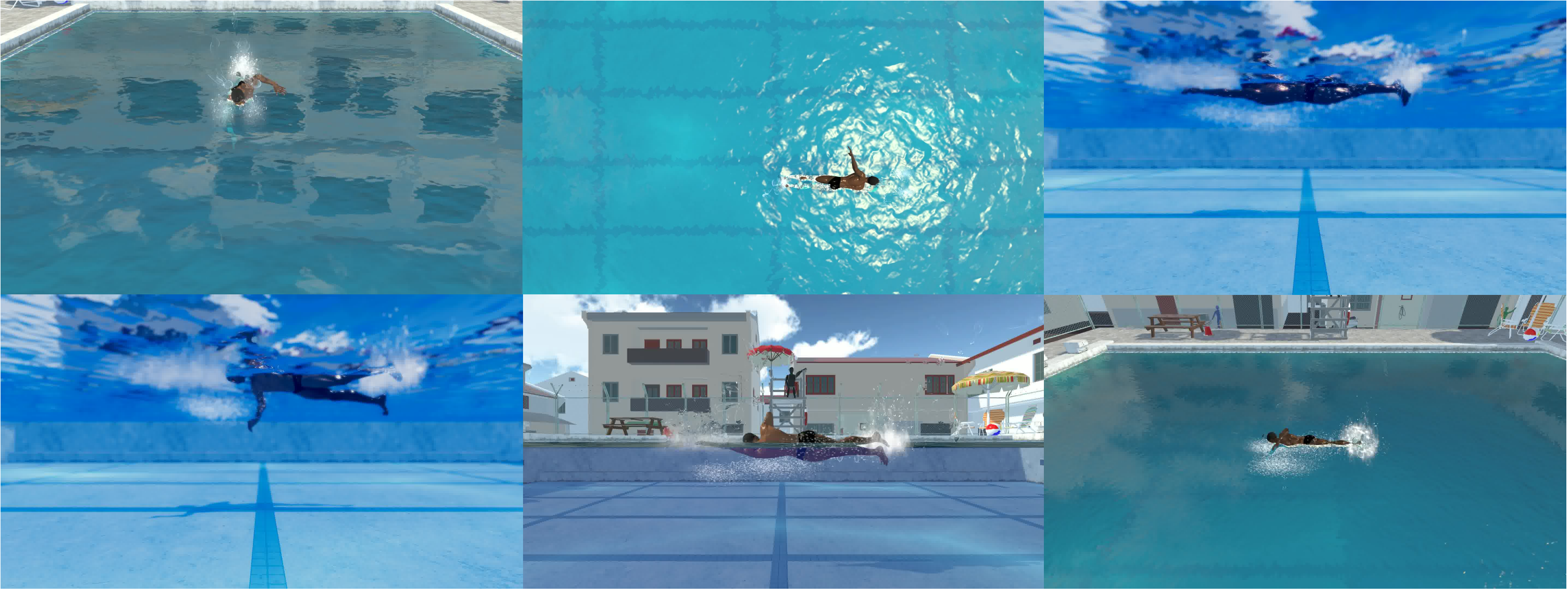}
    \caption{Sample images from SwimXYZ for freestyle motions.}
\end{figure}

\vspace{1em}

\begin{abstract}
Technologies play an increasingly important role in sports and become a real competitive advantage for the athletes who benefit from it. Among them, the use of motion capture is developing in various sports to optimize sporting gestures. Unfortunately, traditional motion capture systems are expensive and constraining. Recently developed computer vision-based approaches also struggle in certain sports, like swimming, due to the aquatic environment. One of the reasons for the gap in performance is the lack of labeled datasets with swimming videos. In an attempt to address this issue, we introduce SwimXYZ, a synthetic dataset of swimming motions and videos. SwimXYZ contains 3.4 million frames annotated with ground truth 2D and 3D joints, as well as 240 sequences of swimming motions in the SMPL parameters format. In addition to making this dataset publicly available, we present use cases for SwimXYZ in swimming stroke clustering and 2D pose estimation. See the project page and download the data at \href{https://g-fiche.github.io/research-pages/swimxyz/}{https://g-fiche.github.io/research-pages/swimxyz/}. 
\end{abstract}

% keywords can be removed

\section{Introduction}\label{intro}
Human motion capture has become a crucial technology in many fields, including character animation for the entertainment industry \cite{Avatar, starke2022deepphase, ling2020character}, medicine \cite{pfister2014comparative, meneguzzo2021body}, and sports \cite{swim, groundReaction, AIcoach}. In sports, Motion Capture is used for multiple applications, such as analyzing and optimizing the athletes' performance \cite{swim, AIcoach}, preventing injuries \cite{touzard2019biomechanical, mauntel2017automated}, animations for the video-game industry \cite{xie2022learning, starke2022deepphase}, or even producing insightful visualization for TV broadcasters \cite{xiu2023econ}. Despite giving reliable results in most scenarios, traditional motion capture systems can hardly be used on a large scale because they are costly and require time-consuming set-up, calibration, and post-processing. These problems are heightened for sports like swimming, raising specific issues due to the aquatic environment, such as reflections on markers or installing underwater cameras. Recent advances have enabled motion capture from RGB images \cite{li2022cliff, kocabas2021pare, kolotouros2019learning, corona2022learned} and videos \cite{kocabas2020vibe} with low-cost and easy-to-use systems. These systems working with only one camera in real-time could pave the way for generalizing the use of motion capture during competitions, taking advantage of already available live video data. Small structures could also use it for optimizing the training of amateur athletes. 

However, we face several limitations in applying computer vision-based motion capture to swimming, all due to a lack of data. Indeed, every Human Pose and Shape (HPS) estimation method needs to extract information from the image, being 2D (2D joints, body segmentation, ...) or 3D (3D joints, virtual markers, ...). Yet, computer-vision models trained on traditional datasets struggle with aquatic data, which are very different from the images they were trained on. Recent works in HPS estimation showed that synthetic data could complement or even replace real images \cite{agora, Black_CVPR_2023, varol17_surreal}.

For generalizing the usage of image-based motion capture methods in swimming, we introduce SwimXYZ. SwimXYZ is a synthetic dataset specialized in swimming, with synthetic monocular videos annotated with ground truth 2D and 3D joints. SwimXYZ consists of 11520 videos for a total of 3.4 million frames with variations in camera angle, subject and water appearances, lighting, and motion. SwimXYZ also features 240 synthetic swimming motion sequences in SMPL format \cite{SMPL:2015}, with variations in body shapes and motion.

\begin{figure*}[!ht]
    \centering
    {\includegraphics[width=\textwidth]{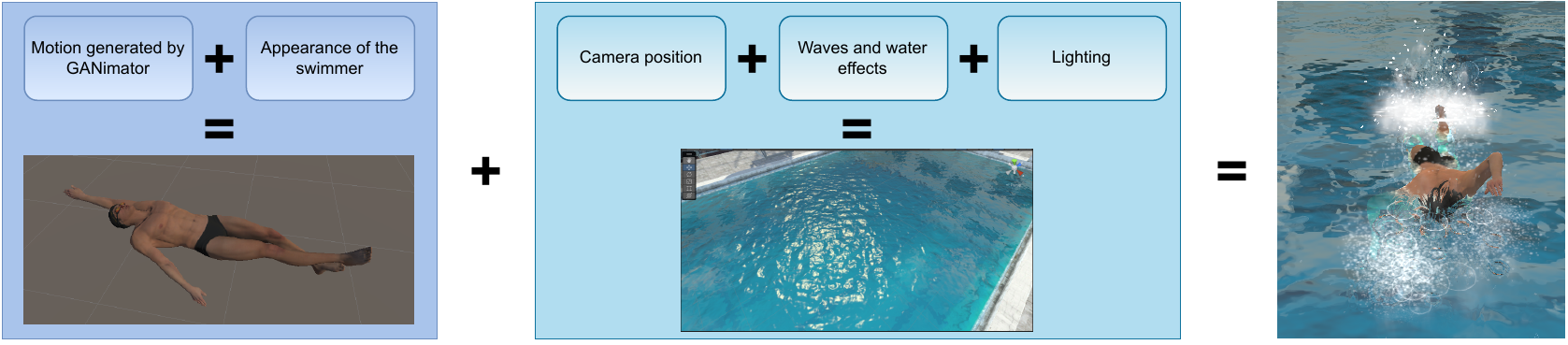}}
    \caption{Pipeline for generating videos. We first generate a unique motion using GANimator \cite{li2022ganimator} trained on a clean swimming motion from \cite{swimmer_unity}, and re-target it on a human body model. We then create an environment by choosing parameters such as camera view, water effects, and lighting. We obtain the final animation by putting the swimmer in the virtual swimming pool.}
    \label{pipeline}
\end{figure*}

\section{Related works}

\subsection{Pose estimation}
\paragraph{3D pose and shape estimation}
While earlier methods only focused on 3D joint locations \cite{martinez_2017_3dbaseline, Fuse2d3d, zhou2016sparseness, OcclusionAware}, more and more works \cite{li2022cliff, kocabas2021pare, kolotouros2019learning, rempe2021humor, tiwari22posendf} regress both pose and body shape. Most approaches are model-based, meaning they estimate the parameters of a parametric human model such as SMPL \cite{SMPL:2015, SMPL-X:2019, xu2020ghum}. Optimization methods use a set of 2D cues \cite{openpose} to iteratively estimate the parameters of the body model given an image \cite{SMPLify, tiwari22posendf} or a video \cite{rempe2021humor, ExploitingTemporalContext}. Most optimization methods also use pose and motion priors \cite{tiwari22posendf, rempe2021humor} to ensure the predictions are realistic. Regression methods use neural networks to regress the parameters of a human body model, given an image \cite{li2022cliff,kocabas2021pare,kolotouros2019learning, hmrKanazawa17} or a video \cite{kocabas2020vibe, humanMotionKanazawa19} as input. While optimization methods usually give the best results, their running time is much longer than regression methods. One of the main difficulties in obtaining better results with regression methods is the lack of data. Recent works proposed several solutions for solving this issue. Hybrid approaches were developed, using optimization procedures to train regression models \cite{kolotouros2019learning} or to generate pseudo-labels for training regression methods in a supervised way \cite{joo2021exemplar}. Another way to train HPS estimation models is to use synthetic data to complement or replace real images \cite{varol17_surreal, Black_CVPR_2023}.

\paragraph{Pose estimation in swimming}
In recent years, several works were published on pose estimation for swimming \cite{ascenso2021development, giulietti2023swimmernet, einfalt2018activity, zecha2017pose}. However, no common benchmark emerged since no dataset was made publicly available. Many of them focused on 2D pose estimation for deriving kinematics parameters such as parts trajectories or stroke rate \cite{zecha2017pose, giulietti2023swimmernet, zecha2012swimmer}. However, most works only used their own videos for training and testing, no dataset was made publicly available, and we have very little information about data diversity. Furthermore, many works were carried out on videos from swimming channels, which can affect the swimmer's motion \cite{einfalt2018activity, zecha2017pose, zecha2012swimmer}. SwimmerNet \cite{giulietti2023swimmernet} was tested on the swimming images of the SVW dataset \cite{sports-videos-in-the-wild-svw-a-video-dataset-for-sports-analysis}, but these images are only labeled with bounding boxes. To our knowledge, even if some works tried performing 3D motion capture with synchronized 2D-annotated video sequences \cite{kirmizibayrak2011digital} or visual hull \cite{ceseracciu2011markerless}, the problem of 3D pose estimation from monocular videos remains unsolved in swimming.

\subsection{Datasets}
\paragraph{Real images}
Datasets of RGB data with accurate 3D pose ground truth annotations are very hard and costly to obtain due to traditional motion capture constraints described in \cref{intro}. Large-scale datasets were collected in a constrained indoor environment, but it is hard to generalize to in-the-wild datasets from these images \cite{ionescu2013human3, Joo_2017_TPAMI}. In-the-wild datasets were also collected, but they are small in scale and lack diversity for training generalizable models \cite{3dpw, bhatnagar22behave}. To alleviate the lack of data, previous approaches rely on datasets annotated with 2D pose for weak supervision \cite{hmrKanazawa17, kocabas2020vibe, kolotouros2019learning}. The main problems with these datasets are that annotations lack consistency due to human annotators and do not provide information about body shape \cite{humanMotionKanazawa19, lin2014microsoft}. Finally, some methods provide pseudo-labels from images \cite{joo2021exemplar, kolotouros2019learning}. These methods are attractive because they can provide low-cost annotations, but these pseudo-labels may lack precision.

\paragraph{Synthetic datasets}
A solution for addressing the lack of high-quality and large-scale datasets is synthetic data. The main advantage of synthetic datasets is that they come with high-quality ground truth labels. They also provide larger diversity, and we can collect as much data as we need once the data generation pipeline has been built. Several works showed the usefulness of synthetic datasets for training HPS estimation models \cite{varol17_surreal, cai2021playing, agora, Black_CVPR_2023}. The most popular method for generating synthetic data is populating a synthetic environment with virtual humans, whose poses and motions are extracted from a real motion capture database \cite{AMASS}. Recently, BEDLAM \cite{Black_CVPR_2023} introduced a dataset enabling a basic HPS estimation model trained only with synthetic data \cite{Black_CVPR_2023, agora} to outperform state-of-the-art models on in-the-wild datasets \cite{li2022cliff, kocabas2021pare}.

\paragraph{Sports datasets}
Motion capture is very useful for sports applications; hence, many datasets with sports images have been published through the years. Most used datasets in sports are LSP \cite{johnson2010clustered} and ASPset \cite{nibali2021aspset}, and many general datasets contain sports images, such as \cite{lin2014microsoft, pennAction}. Datasets were also developed for specific sports, like fitness \cite{tang2023flag3d, fieraru2021aifit, AIcoach}, dance \cite{guo2022multi}, climbing \cite{yan2023cimi4d}, or martial arts \cite{zhang2017martial}. However, to our  knowledge, no dataset for 2D or 3D pose estimation in swimming was made publicly available. \cite{greif2009annotated, ascenso2021development} created datasets for swimming but did not make them available. Recently, \cite{woinoski2020towards} defined criteria for building a database of swimming videos, such as lighting variations, subject diversity, or pools with different depths.

\section{SwimXYZ}

In this section, we present the method for generating the database and detail the different features of SwimXYZ. A general process pipeline is shown in \cref{pipeline}.

\subsection{Preliminaries}\label{preliminaries}
\paragraph{GANimator}
In order to generate diverse swimming motions, we use GANimator \cite{li2022ganimator}, a generative model that learns to synthesize novel motions from a single motion sequence. From a random noise, GANimator learns to produce motions that resemble the core elements of the original motion while simultaneously synthesizing novel and diverse movements. The model is composed of generative and adversarial neural networks \cite{gan}, each responsible for generating motions in a specific frame rate, enabling control of the generated motion across varying levels of detail.

\paragraph{COCO keypoints annotations}
We choose to annotate the video dataset with ground-truth joints in COCO format. COCO \cite{lin2014microsoft} is a large-scale computer-vision dataset, which can be used for tasks such as object detection, segmentation, or human keypoints estimation. COCO has become a reference dataset, and its annotation format is widely used in human pose estimation. Furthermore, foundational models like OpenPose \cite{openpose}, which is used as an off-the-shelf model in many works \cite{SMPLify, rempe2021humor, tiwari22posendf}, predict joints in COCO format. Two different COCO annotation formats exist for keypoints, one with 17 joints and the other with 25 joints.

\paragraph{SMPL model}
Besides being a dataset of videos labeled with 2D and 3D ground-truth joints, SwimXYZ also provides swimming motions in the SMPL \cite{SMPL:2015} format. SMPL is a body model defined as a function $\mathcal{M}(\beta, \theta, \gamma)$ parameterized with the body shape $\beta \in \mathbb{R}^{10}$, the pose $\theta \in \mathbb{R}^{72}$ and the global translation $\gamma \in \mathbb{R}^3$. It outputs 3D joints $J \in \mathbb{R}^{23 \times 3}$ as well as 3D vertices $V \in \mathbb{R}^{6890 \times 3}$ for reconstructing a 3D triangular mesh. In SwimXYZ, we use a more recent version of the SMPL model \cite{MANO}, which only outputs 21 body joints and is parameterized with $\beta \in \mathbb{R}^{16}$, allowing for more detailed body shapes.

\subsection{Generating diverse swimming motions}
In order to generate realistic swimming motions, we started with an existing swimming motion Unity asset \cite{swimmer_unity}. We focused on the swimming part, ignoring the start, and used four animations of this package: backstroke, breaststroke, butterfly, and freestyle. However, \cite{swimmer_unity} only provides a single motion per swimming stroke, which is insufficient to create a database for training deep learning models. To introduce random variations in the swimming motion while keeping it plausible, we used GANimator \cite{li2022ganimator} (see \cref{preliminaries}) trained independently on each swimming stroke. Generated motions are a sequence of 3D joint locations: in order to convert them to SMPL \cite{SMPL:2015} parameters, and before integrating these motions in videos, we re-target generated motions to the 3D swimmer model using Blender \cite{blender} and use the Blender Add-on \cite{SMPL-X:2019} to convert the animation to the SMPL format. Each motion in the database is unique: in addition to variations in swimming strokes, motion is performed at a random speed, from a random position in the swimming pool and towards either end of the basin.

In order to have diversity among the swimmers, different parameters are set randomly: the musculature, the skin color, and the wet skin effect, which affects reflections on the body. The swimmer interacts in the water, with some particle effects representing splashes. Splashes and foam depend on the swimming stroke (there are between 60 and 80 particle effects by swimming stroke), and their intensity is proportional to the speed of the motion.

\subsection{Virtual environment}
The environment was developed using Unity \cite{Unity}. We put much effort into rendering a realistic aquatic environment since the visual difference between aquatic and non-aquatic images is the main difficulty in applying existing models to swimming motions. We have about 50 different parameters for water appearance; however, our developed interface enables setting 12 of them. First, we set the color of the water, with one color for the surface and one for the bottom of the swimming pool. We also set the gradient of the color, which defines how fast the color evolves depending on the depth. Transparency and refraction of the water are also adjustable parameters. Reflection can also be activated with different intensities and distortion effects. We also added waves, for which we set a quantity, a height, and a speed. We set different light sources and colors, which reflect the variation of lighting at different hours of the day. In order to obtain realistic reflections in the water, we added objects and buildings around the pool. We defined five camera views to represent different ways of acquiring data: front, aerial, side underwater, side above water, and side water level.

\subsection{Generation of the database}
\begin{figure}[!t]
    \centering
    \begin{minipage}{0.475\textwidth}
        \begin{minipage}{1\textwidth}
            \resizebox{1.0\linewidth}{!}{\includegraphics{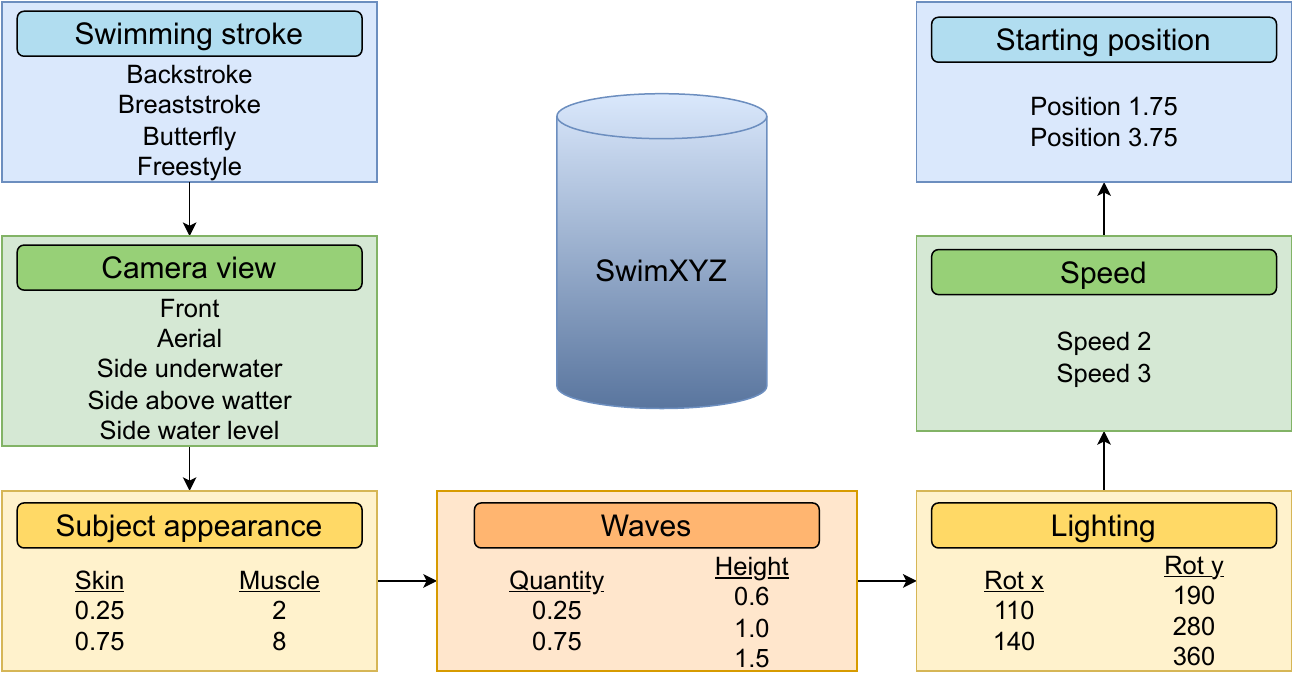}}
            \caption{Structure of the video database.}
            \label{structure}
        \end{minipage}
    \end{minipage}
\end{figure}

We generated the video database using a grid of values for the most important parameters to avoid duplicates. Each video lasts 5 seconds, with 60 images per second frame rate. Videos are annotated in 3D and in 2D, in global coordinates, and in a frame centered on the pelvis of the swimmer. Three annotation formats are available: COCO17, COCO25, and Base (a detailed skeleton with 48 joints). SwimXYZ comprises 11 520 videos for a total of 3.4 million frames annotated in 2D and 3D in three formats and two reference frames. A list of each varying modality is provided in \cref{structure}. We created one video for every possible combination of parameters, and each video shows a unique motion due to variations produced by GANimator \cite{li2022ganimator}. Examples of videos from SwimXYZ are available in the supplementary material.

The database of swimming motions in the SMPL format contains 60 sequences for each of the four swimming strokes, with variations in motion and body shape.

\section{Applications}
\subsection{Swimming stroke clustering}
We evaluate the potential of our database to train a swimming stroke classifier and the variability for each motion in our database. This experiment is performed on the motions in the SMPL parameters format since we do not need visual cues. Hence, our database is composed of 60 motions per swimming stroke. This experiment aims to learn a latent representation of motion and visualize it in a low-dimensional space.

The latent representation of motion is learned using the Periodic Autoencoder (PAE) introduced in DeepPhase \cite{starke2022deepphase}. Based on studies in neurology, this method assumes that any movement can be decomposed into several periodic submovements. Each movement can then be represented in a structured space where each dimension is a periodic function. The PAE extracts a multi-dimensional space from motion, in which it is easier to cluster animations \cite{starke2022deepphase}. The PAE is learned on the joints' velocities; hence, we use the forward pass of the SMPL \cite{SMPL:2015} model to recover the joints from SwimXYZ sequences, and we compute the velocities by finite differences.

\begin{figure}[!t]
    \centering
    \begin{minipage}{0.475\textwidth}
        \begin{minipage}{1\textwidth}
            \resizebox{1.0\linewidth}{!}{\includegraphics{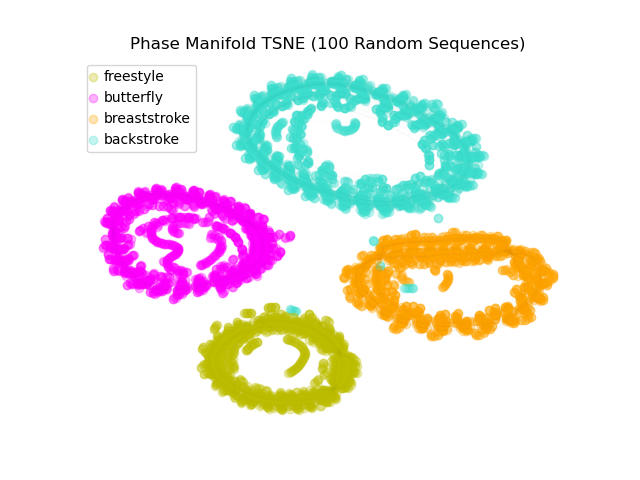}}
            \caption{Visualization of the phase manifold.}
            \label{tsne}
        \end{minipage}
    \end{minipage}
\end{figure}

We train a PAE \cite{starke2022deepphase} with 10 latent channels. A motion sequence of length $T$, $X \in \mathbb{R}^{T \times 21 \times 3}$, is then transformed to a latent representation $X_{lat} \in \mathbb{R}^{T \times 10 \times 2}$ by the PAE. We then visualize 100 random sequences of 2 seconds using the T-SNE algorithm \cite{van2008visualizing}, coloring the sequences depending on the swimming stroke. The results are shown in \cref{tsne}, each point represents a pose, and a complete swimming motion is a circle.

We can see that swimming strokes are almost perfectly separated. Swimming stroke classification being a studied problem in sports science \cite{delhaye2022automatic}, we could imagine training a classifier on SwimXYZ and classifying in-the-wild motions for which we have ground truth or estimated 3D joint positions. One of the main advantages of the SMPL \cite{SMPL:2015} representation is that we can define new joint regressors depending on the desired skeleton layout. We also notice that all motions for the same swimming stroke are not identical since they are not on the exact same circle, showing the diversity of our generated motions.

\subsection{2D pose estimation}

\begin{table}[t]
    \centering
    \caption{Finetuning on SwimXYZ. We compare the finetuned model with the original ViTPose \cite{xu2022vitpose} model in terms of the Percentage of Correct Keypoints with different thresholds given in percentage of the length of the bounding box's diagonal.}
    \begin{tabular}{c|cccc}
                                            & PCK1  & PCK5  & PCK10 & PCK20     \\
    \hline
    ViTPose \cite{xu2022vitpose}            & 13    & 46    & 67    & 87        \\
    ViTPose finetuned                       & 32    & 71    & 85    & 94        \\                    
    \end{tabular}
    \label{quantEval}
\end{table}

\begin{figure*}[!ht]
    \centering
    {\includegraphics[width=\textwidth]{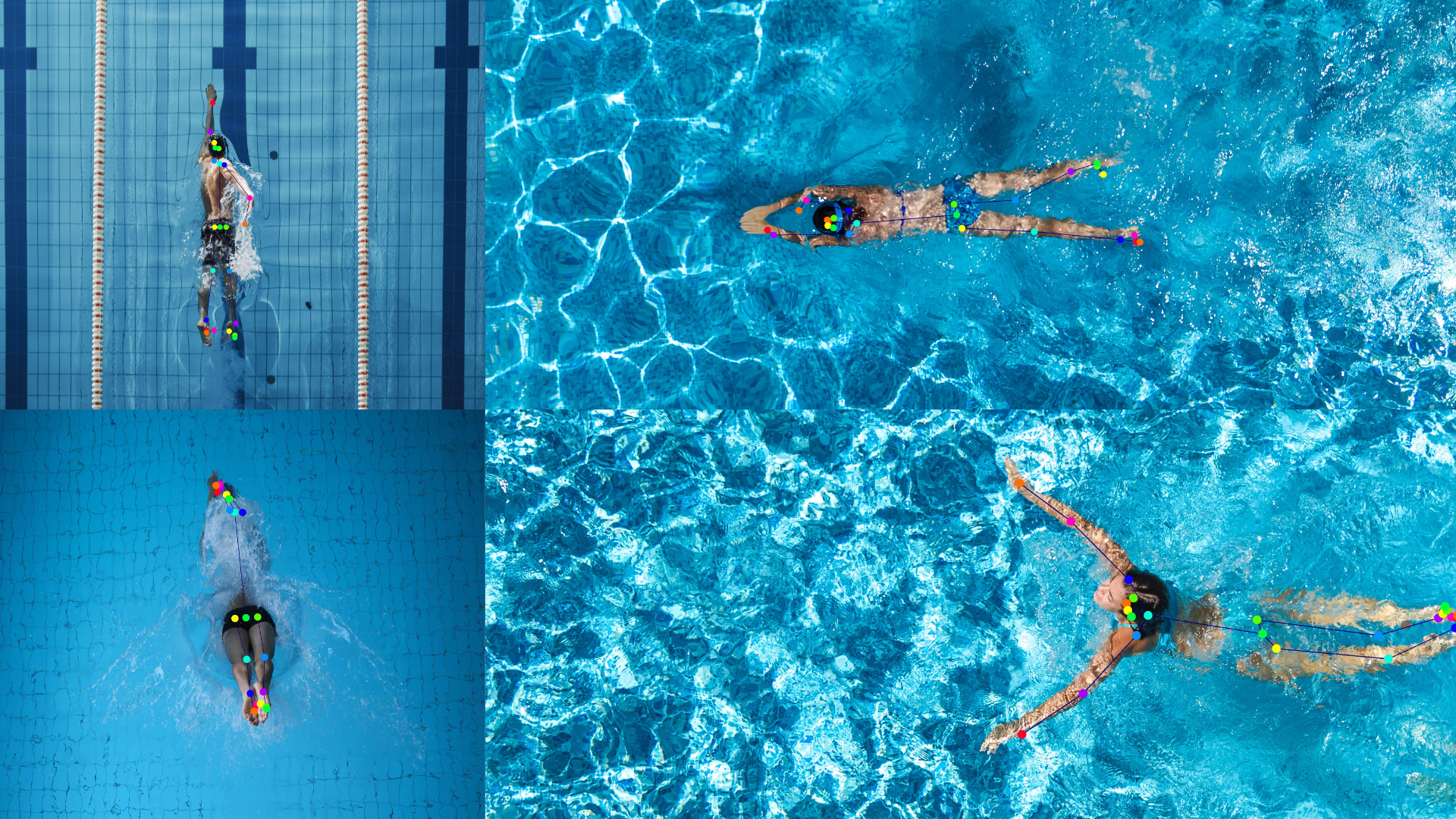}}
    \caption{Qualitative results on real images.}
    \label{qualRes}
\end{figure*}

We also use SwimXYZ for finetuning ViTPose \cite{xu2022vitpose}, the state-of-the-art model for 2D human pose estimation. ViTPose uses vision transformers \cite{dosovitskiy2020image} to extract features from images and a lightweight decoder composed of transposed convolutions and a prediction layer for predicting joints' heatmaps. For this experiment, we choose the ViTPose-L model and finetune both the backbone and the decoder for 5 epochs. The model is trained on a random subset of 5k images of freestyle swimming from SwimXYZ with an aerial view.

First, we evaluate the model qualitatively on real images we collected on the web. Predictions seem satisfactory on most tested images, suggesting that SwimXYZ could be used to train in-the-wild swimming pose estimation systems. Results are shown in \cref{qualRes}.

We also evaluate this model quantitatively on 60 randomly selected synthetic images of butterfly swimming from an aerial view, in order to test a swimming stroke different from the training set. For evaluating models' performance, we use the Percentage of Correct Keypoints (in percent) with different thresholds corresponding to percentages of the size of the bounding box. Results are shown in \cref{quantEval}. We observe that the model finetuned on SwimXYZ data is more accurate than the original model for all tested thresholds. Even for small thresholds, the accuracy is improved by a large margin, showing that finetuning on SwimXYZ enables detecting new keypoints but also improving the precision of detected keypoints.

\section{Conclusion}
In this work, we introduced SwimXYZ, a large-scale dataset of synthetic swimming motions and videos that will be made available online upon acceptance of the paper. SwimXYZ aims to help generalize the use of motion capture in swimming, and our experiments show its potential for future applications. While videos provided by SwimXYZ could be used for training 2D and 3D pose estimation models, future works may use motions in the SMPL format for training pose and motion priors or swimming stroke classifiers. A limitation of SwimXYZ, which may be addressed in future works, is the lack of diversity among subjects (gender, body shape, swimming suit appearance, ...) and environments (surrounding environment, floor of the pool, ...). Other improvements may include other types of annotations (segmentation, depth maps, ...) or adding other parts of the swimming motion, such as dives and turnarounds.

\section{Acknowledgments}
This study is part of the EUR DIGISPORT project supported by the ANR within the framework of the PIA France 2030 (ANR-18-EURE-0022). This work was performed using HPC resources from the “Mésocentre” computing center of CentraleSupélec, École Normale Supérieure Paris-Saclay, and Université Paris-Saclay supported by CNRS and Région Île-de-France. IETR, UMR CNRS 6164 founded the equipment for testing data collection.

%Bibliography
\bibliographystyle{unsrt}  
\bibliography{references}

\end{document}